\documentclass{article} % For LaTeX2e
\usepackage{iclr2024_conference,times}

\usepackage{amsmath,amsfonts,bm}

\def\eqref#1{equation~\ref{#1}}
\def\1{\bm{1}}

\DeclareMathAlphabet{\mathsfit}{\encodingdefault}{\sfdefault}{m}{sl}
\SetMathAlphabet{\mathsfit}{bold}{\encodingdefault}{\sfdefault}{bx}{n}

\usepackage{hyperref}
\usepackage{url}
\usepackage[capitalize]{cleveref}

\usepackage{booktabs} % professional-quality tables
\usepackage{graphicx}
\usepackage{adjustbox} % adjustable tables
\usepackage{dashrule} % table dashed lines

\title{Transformer Neural Autoregressive Flows}

\author{Massimiliano Patacchiola \\
University of Cambridge\\
\texttt{mp2008@cam.ac.uk} \\
\And
Aliaksandra Shysheya \\
University of Cambridge\\
\texttt{as2975@cam.ac.uk} \\
\AND
Katja Hofmann \\
Microsoft Research\\
\texttt{kahofman@microsoft.com} ~~~~~~~~~~~~~~~~~~~~~~~~~~~~~~~~~~~~~~~~~~~~~~~~~~~~~~~~\\
\And
Richard E. Turner \\
University of Cambridge\\
\texttt{ret26@cam.ac.uk} \\
}

\iclrfinalcopy % Uncomment for camera-ready version, but NOT for submission.
\begin{document}

\maketitle

\begin{abstract}
Density estimation, a central problem in machine learning, can be performed using Normalizing Flows (NFs). NFs comprise a sequence of invertible transformations, that turn a complex target distribution into a simple one, by exploiting the change of variables theorem.
Neural Autoregressive Flows (NAFs) and Block Neural Autoregressive Flows (B-NAFs) are arguably the most perfomant members of the NF family. However, they suffer scalability issues and training instability due to the constraints imposed on the network structure.
In this paper, we propose a novel solution to these challenges by exploiting transformers to define a new class of neural flows called Transformer Neural Autoregressive Flows (T-NAFs). T-NAFs treat each dimension of a random variable as a separate input token, using attention masking to enforce an autoregressive constraint. We take an amortization-inspired approach where the transformer outputs the parameters of an invertible transformation. The experimental results demonstrate that T-NAFs consistently match or outperform NAFs and B-NAFs across multiple datasets from the UCI benchmark. Remarkably, T-NAFs achieve these results using an order of magnitude fewer parameters than previous approaches, without composing multiple flows.
\end{abstract}

\section{Introduction}

Normalizing Flows (NFs) have emerged as a powerful paradigm in probabilistic modeling and machine learning, offering a versatile framework for mapping between probability density functions through invertible transformations. NFs are suitable for applications which require models of complex probability distributions that support computationally efficient sampling and density evaluation. Consequently, NPs are useful in density modelling problems and also for variational inference in which they are used to specify approximate posterior distributions.

Most of the research related to NFs focuses on designing expressive transformations that conform to practical constraints. NFs specify an bijective mapping from a simple base distribution to a target distribution so that the transformation of densities formula can be used to specify the density over the target variables. The transformation of densities formula requires the computation of the determinant of the Jacobian of this mapping and so NF models are structured so these objects are simple to compute. For example, one notable class of NFs, known as Autoregressive Flows (AFs), decompose a joint distribution into a product of univariate conditionals. These transformations possess a lower triangular Jacobian, making it computationally tractable to compute their determinants, which is an essential requirement for the application of the change of variables theorem. \cite{kingma2016improved} introduced the concept of inverse autoregressive flows (IAFs), refining AFs by employing a composition of trivially invertible affine transformations to model each conditional, paving the way for more efficient computation while retaining expressiveness. 

Recent advances in the NF landscape include the introduction of Neural Autoregressive Flows (NAFs, \citealt{huang2018neural}) and Block Neural Autoregressive Flows (B-NAFs, \citealt{de2020block}) NAFs and B-NAFs replace the IAF's transformation with a learned bijection implemented via a strictly monotonic neural network. This fundamental paradigm shift makes NAFs and BNAFs more flexible than IAFs, as they are universal approximators of real and continuous distributions. However, despite these advancements, both NAFs and B-NAFs encounter challenges in scalability and efficiency, primarily due to the intricate structures and constraints imposed on their networks. The scalability issue becomes particularly pronounced when dealing with high-dimensional data, where the number of parameters in the network increases significantly. Additionally, the training instability often observed in these models can be attributed to the strict monotonicity and masking constraints, which can limit the expressive power and learning dynamics of the networks. These challenges necessitate a more flexible and scalable approach to enhance the practical applicability of NFs in complex, real-world scenarios.

In response to these limitations, we introduce Transformer Neural Autoregressive Flows (T-NAFs), a novel approach that leverages the power of transformer architectures within the realm of NFs. T-NAFs treat each dimension of a random variable as a separate input token and utilize attention masking to enforce an autoregressive constraint. This design choice allows T-NAFs to handle high-dimensional data more efficiently, as the transformer architecture inherently scales better with dimensionality and does not require the same level of parameterization as traditional feedforward networks. Moreover, T-NAFs offer enhanced flexibility and stability in training by freeing the majority of the parameters from the constraints typically imposed in NAFs and B-NAFs. By incorporating transformers, T-NAFs not only address the scalability and efficiency issues but also open up new avenues for innovation in the field of NFs, particularly in complex density estimation tasks.

The main contributions of this paper can be summarized as follows:
\begin{enumerate}
    \item We introduce Transformer Neural Autoregressive Flows (T-NAFs), a novel model that effectively combines transformers with neural autoregressive flows for density estimation, utilizing attention masking to enforce autoregression efficiently.
    \item We demonstrate remarkable flexibility in handling various invertible transformations and showcase significant improvements in parameter efficiency and scalability compared to previous models like NAFs and B-NAFs.
    \item Through extensive ablation studies and empirical evaluations across multiple benchmarks, we provide key insights into the adaptability, complexity, and effectiveness of the model.
\end{enumerate}

\section{Previous work}

In this section we delve into the extensive body of research surrounding NFs, a domain that has witnessed substantial developments in the last years. For a comprehensive and detailed understanding of the progress in this field, we refer the reader to recent review papers of \cite{kobyzev2019normalizing} and \cite{papamakarios2021nfs}. 

One broad line of research views NFs as a finite number of tractable simple transformations and hence focuses on developing more expressive transformations that have a tractable inverse and Jacobian determinant. \cite{rezende2015vinfs} were the first to introduce two parametric families of such transformations, i.e. the planar and the radial flows. Another well-known representatives of this research direction are Autoregressive Flows (\cref{sec:autoregressive_flows}), which are particularly relevant to our work. \cite{dinh2016density, kingma2016improved} proposed to build Autoregressive Flows with simple affine transformations. To improve the limited expressivity of the affine autoregressive transformations, several non-affine transformations were proposed: \cite{huang2018neural, de2020block} use a monotonic multi-layered perceptron, while \cite{jaini2019integralflow, wehenkel2019monotonicnn} use an integral of some positive function represented as neural network. Although these transformations can be made arbitrary flexible, their major drawback is that they do not have an analytic inverse. To overcome this issue, different spline-based transformations were proposed in \cite{muller2018spline, durkan2019neuralspline, dolatabadi2020lrs}. 

Another important design choice for Autoregressive Flows is how to implement the conditioning on observed variables. Even though there only constraint imposed on conditioning function is the autoregressive one, in practice it is still important to choose a parameter-efficient function that could accommodate inputs of variable size. For instance, \cite{oliva18transformation, kingma2016improved} use recurrent neural network (RNN) as a conditioning function. Even though parameter-efficient, RNNs require sequential computation, which can be prohibitively expensive for longer sequences. As an alternative to RNNs, the idea of masking can be used: usually a standard feedforward network is trained, but some of the connections are zeroed to preserve the autoregressive structure of the conditioning function. \cite{germain2015made} proposed a masking procedure for fully-connected neural networks. Masking was picked up in many autoregressive flows models, including \cite{kingma2016improved, de2020block, huang2018neural, papamakarios2017uci}.

Transformers have been used for distribution modeling in \cite{fakoor2020trade}. The method exploits a transformer encoder that takes the input sequence $x_1, x_2, \dots, x_i$ and predicts the $i$-th conditional distribution $p(x_i | x_1, x_2, \dots, x_{i-1})$. The output is a categorical distribution (discrete variables, or a mixture of Gaussians (continous variables). The main issue with this approach is that it is not possible to get an exact estimate of the likelihood of a sample, as the transformations are not bijective. Moreover, if the input sequence is long the output of the model will be disproportionally large, since each input is mapped to multiple outputs (e.g. categories or parameters of the mixture).

\section{Background}

\subsection{Normalizing Flows}

A Normalizing Flow (NF) is an invertible (bijective) funciton $f: \mathcal{X} \rightarrow \mathcal{Y}$ between two continuous random variables. Since $f$ is invertible, it is possible to use the change of variable theorem to translate between the two densities $p_{X}(\mathbf{x})$ and $p_{Y}(\mathbf{y})$
\begin{equation}\label{eq:change_variable}
    p_{Y}(\mathbf{y}) = p_{X}(\mathbf{x}) \ \big\lvert \text{det} \ \mathbf{J}_{f(\mathbf{x})} \big\rvert^{-1} \quad \text{where} \quad \mathbf{J}_{f(\mathbf{x})} = \frac{\partial \ f(\mathbf{x})}{\partial \ \mathbf{x}}.
\end{equation}
The determinant of the Jacobian $\mathbf{J}_{f(\mathbf{x})}$ accounts for local expansions/contractions of $\mathcal{X}$ at $\mathbf{x}$. Estimating this determinant is one of the major bottlenecks to overcome, since typically this has complexity $\mathcal{O}(N^3)$ for a Jacobian matrix of size $N \times N$. One way to deal with this issue is to enforce a lower-triangular structure on the Jacobian. The determinant of a lower-triangular matrix can be estimated by multiplying the elements on the main diagonal, which has cost $\mathcal{O}(N)$. The lower-triangular structure of the Jacobian can be enforced through an autoregressive factorization, as detailed in the next subsection.

\subsection{Autoregressive Flows} \label{sec:autoregressive_flows} For a multivariate random variable $\mathbf{x} \in \mathbb{R}^{K}$ sampled from $p_{X}(\mathbf{x})$ we can estimate the join distribution with an autoregressive factorization by exploiting the chain rule
\begin{equation}
    p_{X}(\mathbf{x}) = p_{X_1}(x_1) p_{X_2}(x_2 | x_1) \dots p_{X_K}(x_K | x_{K-1}, \dots, x_1).
\end{equation}

Exploiting this structure it is possible to estimate an arbitrary $y_i$ by defining an autoregressive conditioner $c$ and an invertible transformation $t$, such that\begin{equation} \label{eq:t_and_c}
    y_{i} = f(x_1, \dots, x_i) = t( c(x_1, \dots, x_{i-1}), x_i).
\end{equation}

Previous work, has investigated different ways of parametrizing $c$ and $t$. A line of work focused on defining simple affine transformations such as $t(\mu, \sigma, x_i) = \mu + \sigma x_i$ in RealNVP \citep{dinh2016density} or $t(\mu, \sigma, x_i) = \sigma x_i + (1-\sigma) \mu$ in IAF \citep{kingma2016improved}. While those transformations are trivial to invert, they suffer of scarce flexibility which makes them inadequate to model complex distributions. In order to overcome this issue another line of work has proposed to use neural networks to represent the invertible transformation and/or the conditioner. In the next subsection we will describe these particular models more in detail.

\subsection{Neural Autoregressive Flows}

\textbf{Neural Autoregressive Flows} In Neural Autoregressive Flows (NAFs, \citealt{huang2018neural}) Equation~\ref{eq:t_and_c} is reframed as follows
\begin{equation}
    t( c(x_1, \dots, x_{i-1}), x_i) = \text{NN}\big(x_i; \ \boldsymbol{\psi}_i=c(x_1, \dots, x_{i-1}; \ \boldsymbol{\theta}_i) \big)
\end{equation}
where $\text{NN}$ is a feed-forward neural network and $\boldsymbol{\psi}_i$ are pseudo-parameters generated by a conditioner network (parametrized by $\boldsymbol{\theta}$), those pseudo-parameters represent the weights and biases of the neural network. The conditioner network is implemented using a masked feed-forward model as in MADE \citep{germain2015made}, while the invertible transformation is implemented as a deep sigmoidal flow (DSF) or a deep dense sigmoidal flow (DDSF). Both DSF and DDSF exploit sigmoidal inflection points in a fully-connected model to induce multi-modality. A major limitation of NAFs is the use of MADE in the conditioner. With MADE the number of parameters in the network grows with the dimensionality of the random variable, which makes the computational cost prohibitive expensive. Moreover, MADE exploits a zero-masking strategy to impose the autoregressive constraint and to parallelize the computation in the forward pass, which results in a large portion of unused weights that affect the parameter count without having any concrete utility.

\textbf{Block Neural Autoregressive Flows} In order to overcome some of the limitations of NAFs, \cite{de2020block} proposed Block Neural Autoregressive Flows (B-NAFs). The main idea of B-NAF is to directly parametrize the transformation $t$ without the conditioner $c$ by using a single neural network
\begin{equation}
    t( c(x_1, \dots, x_{i-1}), x_i) = \text{NN}\big(x_1, \dots, x_{i-1}; \ \boldsymbol{\theta}_i \big).
\end{equation}
The neural network is a feed-forward model, carefully designed to be both autoregressive and strictly monotone (bijective). The autoregressive constrain is imposed by using masking as in MADE \citep{germain2015made} and the monotonic constraint is imposed by using strictly positive weights and invertible activation functions. While B-NAFs are more effective than NAFs on a variety of datasets \citep{de2020block} they still have two major limitations. The first limitation is that, similarly to NAFs, B-NAFs rely on masked feed-forward networks (MADE) that scale poorly with the dimensionality of the random variable and results in unused parameters due to zero-masking. The second limitation is that enforcing monotonicity in a large model leads to training instability.

In the next section we describe how it is possible to overcome the limitations of NAFs and B-NAFs by exploiting transformers as conditioning mechanisms, presenting a new type of neural autoregressive flow.

\section{Transformer Neural Autoregressive Flows}

In order to overcome the limitations of NAF and B-NAF we propose a new type of flow called Transformer Neural Autoregressive Flow (T-NAF). The main idea of T-NAF is to allocate most of the computational capacity to the conditioner, using a transformer neural network \citep{vaswani2017attention}. This design choice has a series of advantages: (i) we can define an autoregressive constrain by simply using an autoregressive mask in the attention layer, contrary to MADE this does not result in unused weights; (ii) the parameters of the transformer are shared across all the dimensions of the random variable, which significantly reduces the number of total parameters in the model; (iii) the conditioner does not need to be monotonic, which makes training significantly more stable; and (iv) we can exploit all the techniques that have been developed in the last few years to boost the performance of vision and language transformers (e.g. FlashAttention \citealt{dao2022flashattention}, PagedAttention \citealt{kwon2023efficient}).

In T-NAFs we define the invertible transformation and conditioner as follows
\begin{eqnarray} \label{eq:t_and_c_tnaf}
    t( c(x_1, \dots, x_{i-1}), x_i) = t\big(x_i; \ \boldsymbol{\psi}_i=\text{TN}(x_1, \dots, x_{i-1}, i; \ \boldsymbol{\theta}) \big),
    %\\t( c(x_1, \dots, x_{i-1}), x_i) = t_{\phi}\big(x_i; \ \boldsymbol{\psi}_i=\text{TN}_{\theta}(x_1, \dots, x_{i-1}, i) \big),
\end{eqnarray}
where $\text{TN}$ is a Transformer Network parametrized by $\boldsymbol{\theta}$, that takes as input the random variable and the index $i$ (implicitly included via positional encoding) and generates the parameters $\boldsymbol{\psi}_i$ of the transformation $t$. We do not impose any particular constraint on the transformation $t$, which can be any type of invertible transformation. In Section~\ref{sec:transformation} we describe in more details some of the transformations we have used in our experiments. Empirically we have observed that a compact monotonic neural network (taking the variable $x_i$ as input and producing $y_i \in [0,1]$ as output) offers the best trade-off between performance and model overhead. This is equivalent to a parametric Cumulative Distribution Function (CDF). Crucially, because of the single input-output structure, this network does not require MADE masking and it is stable to train due to its limited size.

Importantly, in Equation~\ref{eq:t_and_c_tnaf} there is a single set of parameters $\boldsymbol{\theta}$ in the conditioner that are shared across all the dimensions of the random variable, while in both NAFs and B-NAFs there is a separate set of parameters $\boldsymbol{\theta}_i$ for each dimension $x_i$. This means that T-NAFs requires significantly less parameters (see the experimental section for more details). 

\begin{figure}[t]
\centering
\includegraphics[width=\textwidth]{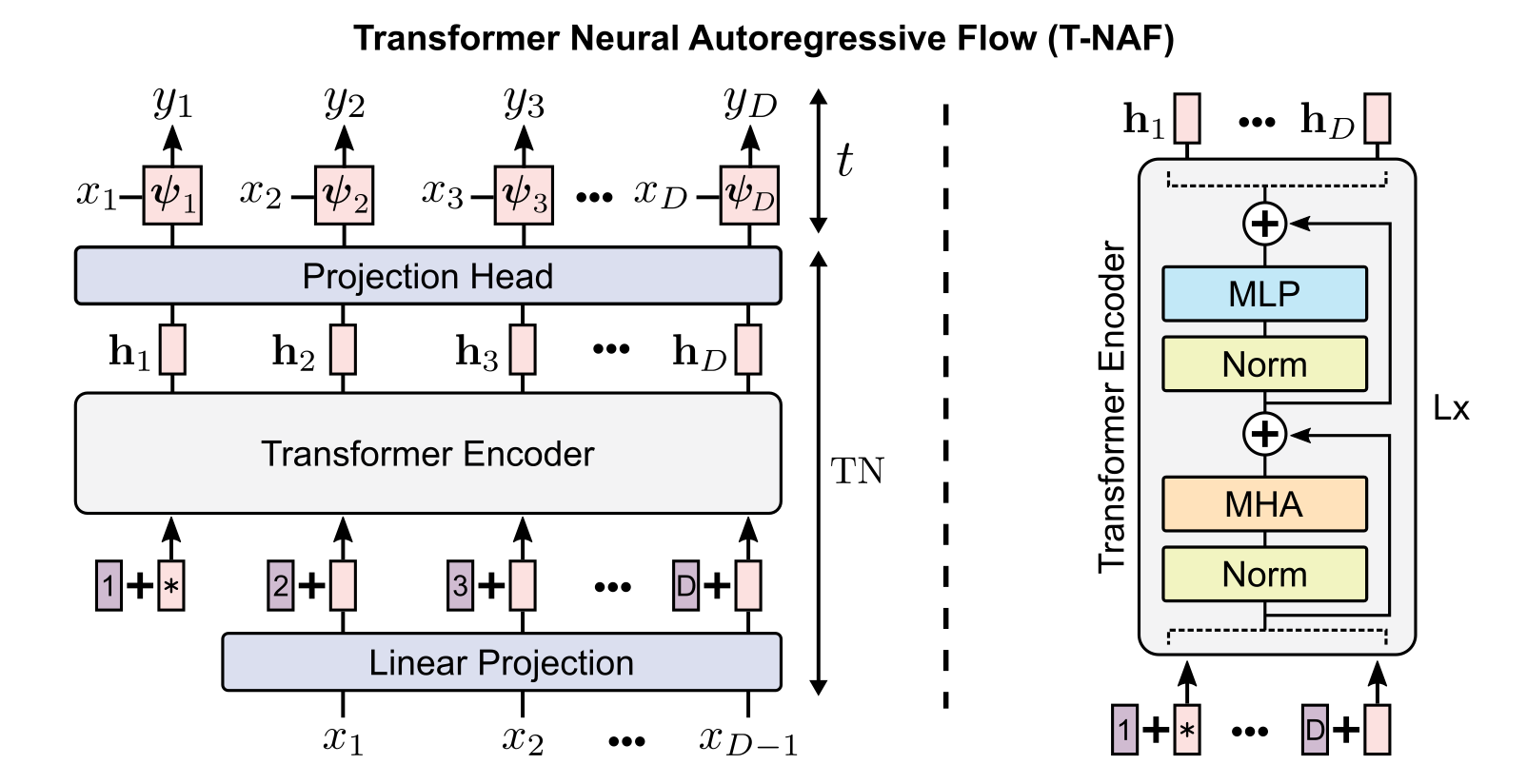}
\caption{Graphical representation of a T-NAF model. \textbf{Left}: the architecture includes a transformer neural network conditioner $\text{TN}$ and an invertible transformation $t$. In $\text{TN}(x_1, \dots, x_D, i; \boldsymbol{\theta})$ each dimension of the random variable is linearly projected to get an embedding which is added to a learnable position vector. A sequence of $L$ transformer layers is used to produces hidden embeddings $\mathbf{h}_1, \dots, \mathbf{h}_D$ that are passed through a projection head to generate pseudo-parameters $\boldsymbol{\psi}_1, \dots, \boldsymbol{\psi}_D$. The pseudo-parameters are used as part of an invertible transformation $t(x_i; \boldsymbol{\psi}_i)$. \textbf{Right}: detailed schematic of a transformer encoder layer. Inputs are passed through a normalization layer, a Multy Head Attention (MHA) layer with autoregressive mask, another normalization layer, and an MLP.}
\label{fig:tnaf_schematic}
\end{figure}

\subsection{Overview of the architecture} 

The overall design of the proposed transformer is similar to the Vision Transformer (ViT, \citealt{dosovitskiy2020image}) with two key differences: (i) we replace the convolutional projections with simple linear projections, as we are not dealing with image patches; (ii) we replace the classification head (taking as input only the first embedding in the sequence) with a projection head that generates the pseudo-parameters (for all the embeddings in the sequence). Figure~\ref{fig:tnaf_schematic} showcase a high-level overview of the T-NAF architecture.

In the initial stage of the pipeline, each dimension of the random variable is linearly projected to get inputs embeddings, those are added to learnable positional embeddings and passed to a series of $L$ transformer encoder layers. The first element of the autoregressive factorization is the marginal distribution $p(x_1)$, which does not require any conditioning. To impose this constraint we use a learnable input embedding as first element of the sequence. This guarantees that the first set of pseudo-parameters $\boldsymbol{\psi}_1$ is generated without having access to any element of the sequence. This is similar in spirit to the Beginning of Sequence (BoS) token or class token used in a variety of transformer models \citep{devlin2018bert, dosovitskiy2020image}.

A detailed representation of the encoder layer is provided on the right-hand side of Figure~\ref{fig:tnaf_schematic}. The encoder layer performs the following operations: normalization via layer-norm, attention estimation via Multi-Head Attention (MHA), normalization via layer-norm, forward pass in a Multi-Layer Perceptron (MLP). Skip connections are used from the input to the output of the MHA, and from the output of the MHA to the output of the encoder.
An autoregressive mask is used in the MHA to enforce the autoregressive constraint. The output of the encoder is a series of hidden embedding vectors $\mathbf{h}_i$, which are passed to a projection head to generate the pseudo-parameters $\boldsymbol{\psi}_i$. The pseudo-parameters are used to define the invertible transformation $y_i = t(x_i; \ \boldsymbol{\psi}_i)$. The projection head can be a set of linear layers that takes as input the embeddings $\mathbf{h}_i$ and generates the pseudo-parameters $\boldsymbol{\psi}_i$, or it can simply be an identity function (see Section~\ref{sec:transformation} for additional details).

\subsection{Transformation} \label{sec:transformation}
The modularity of the architecture allows using a large and diversified set of transformations, below we detail some of them.

\textbf{Affine transformation} One of the simplest transformation is the affine one. An affine transformation is defined by a scale $\mu_i$ and shift $\sigma_i$ parameter, grouped into a set of pseudo-parameters $\boldsymbol{\psi}_i = \{\mu_i, \sigma_i \}$, that are obtained as follows
\begin{equation} \label{eq:t_affine}
    t_{\text{affine}}(x_i; \ \boldsymbol{\psi}_i) = \mu_i + \sigma_i \ x_i
    \quad \text{with} \quad \mu_i=\text{LIN}_{\mu}(\mathbf{h}_i; \boldsymbol{\theta}_{\mu}), \quad \sigma_i=\text{LIN}_{\sigma}(\mathbf{h}_i; \boldsymbol{\theta}_{\sigma}),
\end{equation}
where $\text{LIN}_{\mu}$ and $\text{LIN}_{\sigma}$ are linear layers parametrized by $\boldsymbol{\theta}_{\mu}$ and $\boldsymbol{\theta}_{\sigma}$ used to generate the scale $\mu_i$ and shift $\sigma_i$. The linear layers of the projection head are shared across all the output tokens. Note that, it is possible to concatenate multiple affine transformations via function composition, in this case a separate set of linear layers can be assigned to each transformation.

\textbf{CDF transformation} A more sophisticated transformation is given by a parametric Conditional Distribution Function (CDF) that maps $x_i$ to a uniform distribution. Following previous work, we can define the parametric CDF as a neural network with positive weights \citep{archer1993application, sill1997monotonic, daniels2010monotone, de2020block}. 
In this case the pseudo-parameters are the weights and biases of the neural network $\boldsymbol{\psi}_i = \{\mathbf{w}^{(1)}_i, \mathbf{b}^{(1)}_i, \mathbf{w}^{(2)}_i, b^{(2)}_i \}$ which are used as part of the following transformation
\begin{equation} \label{eq:t_cdf}
    t_{\text{CDF}}(x_i; \ \boldsymbol{\psi}_i) =  \text{sig}\left( \text{tanh}\left( \exp \mathbf{w}^{(1)}_{i} \ x_i + \mathbf{b}^{(1)}_i \right)^{T} \exp \mathbf{w}^{(2)}_i + b^{(2)}_i \right),
\end{equation}
where $\text{sig}(\cdot)$ and $\text{tanh}(\cdot)$ are the sigmoid and hyperbolic tangent functions. Similarly to Equation~\ref{eq:t_affine}, the weights and biases are generated via dedicated linear layers $\mathbf{w}^{j}_i=\text{LIN}_{\mathbf{w}^{j}}(\mathbf{h}_i; \boldsymbol{\theta}_{\mathbf{w}^{j}})$ and $\mathbf{b}^{j}_i=\text{LIN}_{\mathbf{b}^{j}}(\mathbf{h}_i; \boldsymbol{\theta}_{\mathbf{b}^{j}})$ as part of the projection head. The exponential function used in Equation~\ref{eq:t_cdf} ensures that the weights are strictly-positive enforcing the monotonic constraint (softplus can also be used), this and the use of invertible activation functions makes the transformation bijective. It is possible to invert $t_{\text{CDF}}$ since it has a single input, a single output, and is bijective. The inverse of $t_{\text{CDF}}$ is an inverse-CDF, which can be used for inversion sampling following two steps: (i) first sample a value $y_i$ from the uniform distribution, (ii) find the corresponding $x_i$ via a root-finding method (e.g. the bisection method).

\textbf{Shared-CDF transformation} In the CDF transformation defined in the previous paragraph, the pseudo-paramters are generated by using the projection head and the transformer embeddings. An alternative approach is to define the CDF neural network in advance, using a fixed set of weights that are shared across all the outputs. In this case, $y_i$ is obtained via a conditional CDF, that takes as input $x_i$ and the transformer embeddings $\mathbf{h}_i$. Assuming that the conditional CDF is modeled using a neural network with a single hidden layer, we can formalize this particular case defining a set of pseudo-parameters $\boldsymbol{\psi}_i$ and a set of shared-parameters $\boldsymbol{\phi}$:
\begin{equation}
    \boldsymbol{\psi}_i = \left\{ \mathbf{h}_i \right\} \quad \text{and} \quad
    \boldsymbol{\phi} = \left\{\mathbf{w}^{(1)}, \mathbf{w}^{(2)}, \mathbf{b}^{(1)}, b^{(2)}, \mathbf{\hat{W}}^{(1)}, \mathbf{\hat{w}}^{(2)}\right\}.
\end{equation}
The pseudo-parameters coincide with the embeddings produced by the transformer conditioner $\mathbf{h}_i = \text{TN}\left( x_1, \dots, x_{i-1}, i; \boldsymbol{\theta} \right)$.
The shared-parameters include $\mathbf{w}^{(1)}, \mathbf{w}^{(2)}, \mathbf{b}^{(1)}, b^{(2)}$ which are related to the input $x_i$, and $\mathbf{\hat{W}}^{(1)}, \mathbf{\hat{w}}^{(2)}$ which are related to the conditional input $\mathbf{h}_i$. 
The shared-CDF transformation is defined as follows
\begin{equation} \label{eq:shared_cdf}
    t_{\text{S-CDF}}(x_i; \ \boldsymbol{\psi}_i, \boldsymbol{\phi}) =  \text{sig}\left( \text{tanh}\left( \exp \mathbf{w}^{(1)} \ x_i + \mathbf{\hat{W}}^{(1)} \ \mathbf{h}_i + \mathbf{b}^{(1)} \right)^{T} \exp \mathbf{w}^{(2)} + \mathbf{\hat{w}}^{(2)} \mathbf{h}_i + b^{(2)} \right),
\end{equation}
where we have discarded any reference to the projection head, as we can assume it corresponds to the identity function. Note the crucial distinction between positive-constraint weights $\exp \mathbf{w}^{(1)}, \exp \mathbf{w}^{(2)}$ used for the input and the unconstrained weights $\mathbf{\hat{W}}^{(1)}, \mathbf{\hat{w}}^{(2)}$ used in the conditioning mechanism. The main advantage of the shared-CDF approach is that the transformer needs to generate just the vector of embeddings $\mathbf{h}_i$ for each token, while in the CDF transformation it needs to generate the weights and biases of the entire CDF network. Therefore, it is more convenient to use a shared-CDF transformation when the size of the CDF neural network is large.

\textbf{Spline transformation} Following \citealt{durkan2019neuralspline}, we also incorporated a spline transformation based on monotonic rational-quadratic splines. Besides being comparable to the CDF transformation in terms of performance and capacity~\citep{durkan2019neuralspline}, the spline transformation has the advantage of being easily invertible. Each block of this transformation consists of a elementwise rational-quadratic spline
\begin{equation}
    t_{\text{spline}}(x_i; \ \boldsymbol{\psi}_i) = g(x_i; \boldsymbol{\psi}_i = \text{TN}\left( x_1, \dots, x_{i-1}, i; \boldsymbol{\theta} \right))
\end{equation}
where $g(\cdot; \ \boldsymbol{\psi})$ is a monotonic rational-quadratic spline function parameterized by a set of pseudo-parameters $\boldsymbol{\psi}$ generated by the transformer conditioner (see \citealp{durkan2019neuralspline} for the details on the parameterization of the spline). To ensure cross-variable interactions, the elementwise function is commonly followed by an invertible linear transformation across variables
\begin{equation}
t_{\text{linear}}(\mathbf{x}; \ \boldsymbol{\phi}) = \mathbf{P}\mathbf{L}\mathbf{U}\mathbf{x}
\end{equation}
with the set of shared-parameters $\boldsymbol{\phi} = \{\mathbf{P}, \mathbf{L}, \mathbf{U}\}$, where $\mathbf{P}$ is a fixed permutation matrix, $\mathbf{L}$ is lower triangular with ones on the diagonal, and $\mathbf{U}$ is upper triangular. We generate a set of $\boldsymbol{\psi}_i^j$ from the initial variables $x_1, \dots, x_{i-1}$ at each spline block transformation $j$. However, this means that the suggested $t_{\text{linear}}(\mathbf{x}; \ \boldsymbol{\phi})$ breaks the autoregressive constraint and the determinant of the Jacobian cannot be computed in an efficient way. To fix this issue, we simplify the linear transformation to 
\begin{equation}
    t_{\text{linear}}(\mathbf{x}; \ \boldsymbol{\phi}) = \mathbf{L}\mathbf{x}
\end{equation}
with $\boldsymbol{\phi} = \{\mathbf{L}\}$ for all spline transformation blocks except for the first one. As a result, each block of spline transformation is a composition of $t_{\text{spline}}$ and $t_{\text{linear}}$, where $t_{\text{spline}}$ is applied elementwise and $t_{\text{linear}}$ acts on the whole set of variables.

\section{Experiments}

\subsection{Density Estimation}

In this section we evaluate the performance of our method on density estimation with real-world data, using classic benchmarks from the UCI repository. Following standard practice, we evaluate our model on four datasets from UCI (POWER, GAS, HEPMASS, MINIBOONE) and on BSDS300 \citep{martin2001database} a dataset obtained by extracting random patches from the homonym datasets of natural image. We compare against a variety of methods: RealNVP~\citep{dinh2016density}, Glow~\citep{kingma2018glow}, MADE~\citep{germain2015made}, MAF~\citep{papamakarios2017uci}, TAN~\citep{oliva18transformation}, FFORJD~\citep{grathwohl2018ffjord}, SOS~\citep{jaini2019integralflow}, RQ-NSF~\citep{durkan2019neuralspline}. We also include neural autoregressive models such as NAF~\citep{huang2018neural} and B-NAF~\citep{de2020block}. The results for other methods are taken from the corresponding papers. Note that some methods (FFJORD, B-NAF and TAN) perform a hyper-parameter search over each dataset while we used the same hyper-parameters in all experiments. We used the following configuration for T-NAF: one flow, three or five layers in the transformer encoder, embeddings of size 32, eight attention heads, one hidden layer (64 units) in the encoder MLPs, the projection head generates the parameters of a CDF network with one hidden layer (128 units).

\textbf{Results} The results are reported in Table~\ref{tab:uci}. Overall, T-NAF is able to outperform all the other methods on HEPMASS using the larger model with five layers. On BSDS300 and GAS T-NAF obtained the second highest score after TAN and RQ-NSF, but without performing any ad-hoc hyper-parameter search and without using multiple flows. The only dataset where other methods outperform T-NAF is MINIBOONE.
T-NAF is more efficient than other models for two reasons: a reduced number of flows and reduced number of model parameters. In terms of number of flows, most of the other methods concatenate multiple transformations and permute the order of the variable in between. While this attenuates possible issues due to the variable ordering, it comes at the price of an increased overhead and parameter count. T-NAF achieves top performance using only one flow, therefore substantially reducing the complexity of the model.

\begin{table}[t]
  \caption{Log-Likelihood (higher is better) for 4 datasets from UCI and for BSDS300. Average and standard deviation over 3 seeds. We report number of dimensions ($D$) and dataset size ($N$, Millions). T-NAF obtains some of the best results by using just one flow and no ad hoc architectures. $^{\ast}$Methods using ad hoc architectures on each dataset. Best result in bold, second-best underlined.
  }
  \centering
  \begin{adjustbox}{width=1.0\textwidth}
  \begin{tabular}{lcccccc}
    \toprule
      & & POWER & GAS & HEPMASS & MINIBOONE & BSDS300 \\
      & & $D=6$ & $D=8$ & $D=21$ & $D=43$ & $D=63$ \\
      Model & Flows & $N=2.0$ & $N=1.0$ & $N=0.5$ & $N=0.04$ & $N=1.3$ \\
    \midrule
    RealNVP & 5 & -0.02$\pm$\small{0.01} & 4.78$\pm$\small{1.80}  & -19.62$\pm$\small{0.02} & -13.55$\pm$\small{0.49} & 152.97$\pm$\small{0.28} \\
    RealNVP & 10 & 0.17$\pm$\small{0.01} & 8.33$\pm$\small{0.14}  & -18.71$\pm$\small{0.02} & -13.84$\pm$\small{0.52} & 153.28$\pm$\small{1.78} \\
    Glow & n/a & 0.17$\pm$\small{0.01} & 8.15$\pm$\small{0.40}  & -18.92$\pm$\small{0.08} & -11.35$\pm$\small{0.07} & 155.07$\pm$\small{0.03} \\
    MADE MoG & n/a & 0.40$\pm$\small{0.01} & 8.47$\pm$\small{0.02}  & -15.15$\pm$\small{0.02} & -12.27$\pm$\small{0.47} & 153.71$\pm$\small{0.28} \\
    MAF Affine & 5 & 0.14$\pm$\small{0.01} & 9.07$\pm$\small{0.02}  & -17.70$\pm$\small{0.02} & -11.75$\pm$\small{0.44} & 155.69$\pm$\small{0.28} \\
    MAF Affine & 10 & 0.24$\pm$\small{0.01} & 10.08$\pm$\small{0.02}  & -17.73$\pm$\small{0.02} & -12.24$\pm$\small{0.45} & 154.93$\pm$\small{0.28} \\
    MAF MoG & 5 & 0.30$\pm$\small{0.01} & 9.59$\pm$\small{0.02}  & -17.39$\pm$\small{0.02} & -11.68$\pm$\small{0.44} & 156.36$\pm$\small{0.28} \\
    FFJORD$^{\ast}$ & n/a & 0.46$\pm$\small{0.01} & 8.59$\pm$\small{0.12}  & -14.92$\pm$\small{0.08} & -10.43$\pm$\small{0.04} & 157.40$\pm$\small{0.19} \\
    SOS & 7 & 0.60$\pm$\small{0.01} & 11.99$\pm$\small{0.41}  & -15.15$\pm$\small{0.10} & -\underline{8.90}$\pm$\small{0.11} & 157.48$\pm$\small{0.41} \\
    NAF-DDSF & 5 & 0.62$\pm$\small{0.01} & 11.91$\pm$\small{0.13}  & -15.09$\pm$\small{0.40} & -\textbf{8.86}$\pm$\small{0.15} & 157.73$\pm$\small{0.04} \\
    NAF-DDSF & 10 & 0.60$\pm$\small{0.02} & 11.96$\pm$\small{0.33} & -15.32$\pm$\small{0.23} & -9.01$\pm$\small{0.01} & 157.43$\pm$\small{0.30} \\
    B-NAF$^{\ast}$ & 5 & 0.61$\pm$\small{0.01} & 12.06$\pm$\small{0.02}  & -14.71$\pm$\small{0.38} & -8.95$\pm$\small{0.07} & 157.36$\pm$\small{0.03} \\
    TAN$^{\ast}$ & 5 & 0.60$\pm$\small{0.01} & 12.06$\pm$\small{0.02}  & -\underline{13.78}$\pm$\small{0.02} & -11.01$\pm$\small{0.48} & \textbf{159.80}$\pm$\small{0.07} \\
    RQ-NSF (AR) & 10 & \textbf{0.66}$\pm$\small{0.01} & \textbf{13.09}$\pm$\small{0.02} & -14.01$\pm$\small{0.03} & -9.22$\pm$\small{0.48} & 157.31$\pm$\small{0.28} \\
    \textbf{T-NAF/3 (ours)} & \textbf{1} & \underline{0.63}$\pm$\small{0.01} & 12.01$\pm$\small{0.02}  & -14.88$\pm$\small{0.37} & -11.59$\pm$\small{0.67} & 157.83$\pm$\small{0.10} \\
    \textbf{T-NAF/5 (ours)} & \textbf{1} & 0.54$\pm$\small{0.01} & \underline{12.27}$\pm$\small{0.01}  & -\textbf{13.20}$\pm$\small{0.26} & -10.67$\pm$\small{0.06} & \underline{159.41}$\pm$\small{0.04} \\
    \bottomrule
  \end{tabular}
  \end{adjustbox}
  \label{tab:uci}
\end{table}

\begin{figure}[t]
\centering
\includegraphics[width=\textwidth]{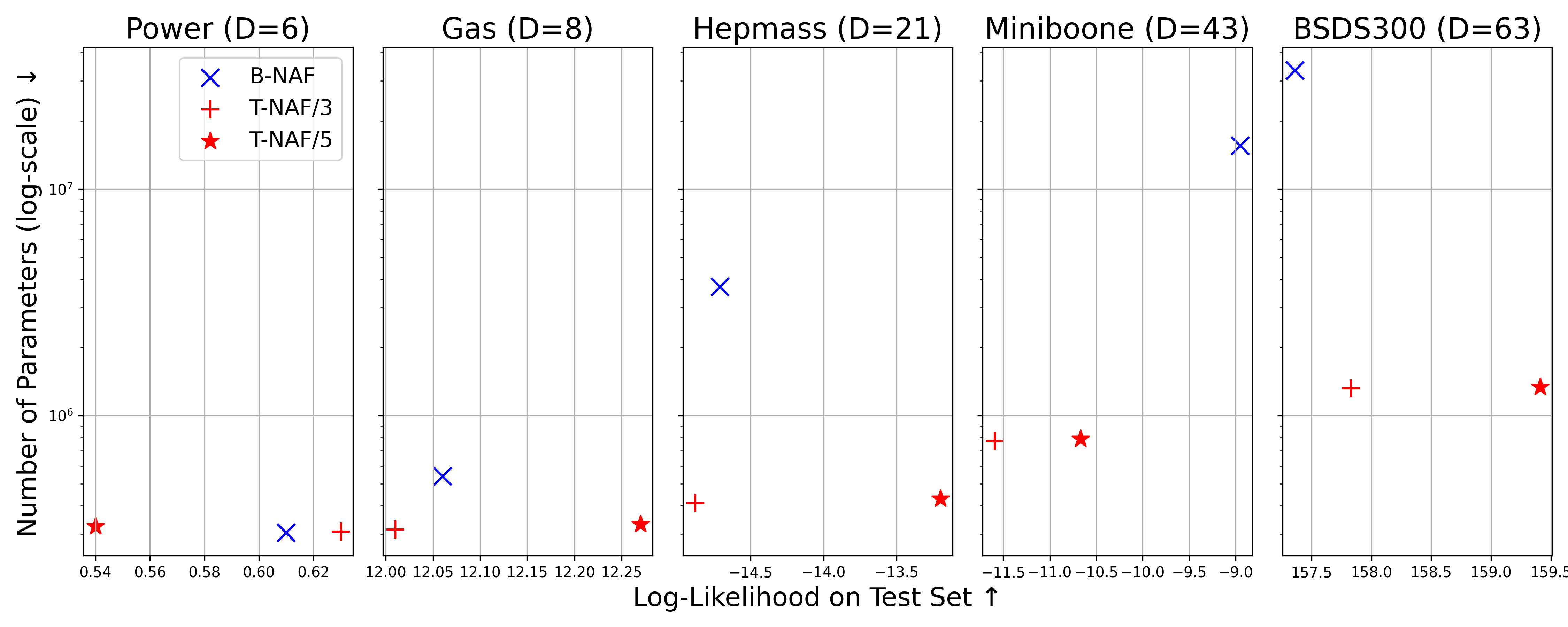}
\caption{Trade-off between number of parameters and performance. Vertical axis represents the number of parameters in log-scale (lower is better), and horizontal axis represent the log-likelihood on the test set (higher is better). The optimal trade-off is represented by points in the bottom-right corner. Overall T-NAF offers a better trade-off w.r.t.~B-NAF; the gap in number of parameters between T-NAF and B-NAF gets larger as the number of input dimensions $D$ increases. }
\label{fig:parameters_vs_performance}
\end{figure}

\textbf{Parameter efficiency} We compared T-NAF and B-NAF in terms of number of parameters vs.~test log-likelihood on the UCI and BSDS300 datasets. By assessing the total parameter count required by each model, we aim to demonstrate how T-NAFs optimize parameter usage. This evaluation offers insights into the practical implications of employing transformer-based architectures in probabilistic modeling, especially in scenarios where computational resources and model simplicity are important. For B-NAF we report the parameters and performance of the best model as specified in \cite{de2020block}, that is: MLP with two hidden layers, number of hidden units defined as $D \times 40$, and five flows. For our T-NAF model we report the number of parameters and performance for the variant based on a CDF transformation and a transformer with 3 and 5 layers. For a fair comparison we also include the T-NAF pseudo-parameters $\boldsymbol{\psi}_1, \dots, \boldsymbol{\psi}_{D}$ generated during the inference step.
The results are summarized in Figure~\ref{fig:parameters_vs_performance}. The figure shows that T-NAF offers a better trade-off in terms of parameter count vs.~performance. In particular, the gap between the two methods gets larger as the dimensionality of the input increases.

\subsection{Ablations} 

In this section, we provide a detailed ablation study to investigate the impact of different components of our T-NAF on performance. A critical aspect of T-NAF is the use of different invertible transformations and their effects on the model's density estimation capabilities. Our experiments primarily focus on contrasting the performance of three distinct types of transformations: Conditional Distribution Function (CDF), shared CDF, and Spline transformations. Each of these transformations offers unique characteristics: CDF transformations are expressive and allow mapping inputs to a uniform distribution, potentially capturing complex patterns in the data more effectively; Spline transformations can offer high levels of flexibility and are expected to model intricate data distributions with greater precision. Additionally, the experiments compare the Shared-CDF approach, where a fixed set of weights is shared across outputs, with the standard CDF approach, where the transformer generates the parameters of the neural-CDF separately for each output. This comparison is crucial to evaluate the benefits of parameter sharing versus individual parameterization in the context of density estimation tasks.
Furthermore, we explore the impact of varying the number of layers in the transformer encoder on the overall performance of the model. This exploration helps in understanding the trade-off between model complexity (in terms of depth) and performance gains. A deeper model might capture more complex relationships in the data but at the cost of increased computational requirements and potential overfitting.

\textbf{Results} A summary of the results is presented in Table~\ref{tab:uci_ablations}. 
The standard CDF model with five layers consistently outperforms the remaining models across most datasets. The Spline Flow with five layers is competitive on three datasets (POWER, HEPMASS, and BSDS300) while suffering the most on MINIBOONE.
Overall, the increase in transformer layers from three to five show a noticeable improvement across the board. This outcome indicates that additional layers might help capture more complex dependencies. Only in one case additional layers seems to reduce the performance (POWER dataset). Using a shared CDF over the standard one does not seem to give particular benefits, there is only a marginal advantage on MINIBOONE and BSDS300 when compared with the CDF with three layers.

\begin{table}[t]
  \caption{Ablation experiments on head type. CDF: the parameters of a CDF are generated by the transformer for each output separately; Shared-CDF: conditional CDF with parameters shared across outputs and conditional input provided by the transformer embeddings;  Spline Flow: the parameters of a spline flow are generated by the transformer for each output separately. 
  Log-Likelihood (higher is better) for 4 datasets from UCI and for BSDS300. Average and standard deviation over 3 seeds. We report number of dimensions ($D$) and dataset size ($N$, Millions).
  }
  \centering
  \begin{adjustbox}{width=1.0\textwidth}
  \begin{tabular}{lcccccc}
    \toprule
      & & POWER & GAS & HEPMASS & MINIBOONE & BSDS300 \\
      & & $D=6$ & $D=8$ & $D=21$ & $D=43$ & $D=63$ \\
      Head Type & Layers & $N=2.0$ & $N=1.0$ & $N=0.5$ & $N=0.04$ & $N=1.3$ \\
    \midrule
    CDF & 3 & 0.63$\pm$\small{0.01} & 12.01$\pm$\small{0.02}  & -14.88$\pm$\small{0.37} & -11.59$\pm$\small{0.67} & 157.83$\pm$\small{0.10} \\
    Shared-CDF & 3 & 0.52$\pm$\small{0.02} & 11.12$\pm$\small{0.09} & -14.97$\pm$\small{0.17} & -11.34$\pm$\small{0.03} & 158.10$\pm$\small{0.07} \\
    Spline Flow & 3 & \textbf{0.64}$\pm$\small{0.01} & 11.99$\pm$\small{0.03}  & -14.05$\pm$\small{0.14} & -11.08$\pm$\small{0.04} & 159.05$\pm$\small{0.07} \\
    CDF & 5 & 0.54$\pm$\small{0.01} & \textbf{12.27}$\pm$\small{0.01}  & -\textbf{13.20}$\pm$\small{0.26} & \textbf{-10.67}$\pm$\small{0.06} & \textbf{159.41}$\pm$\small{0.04} \\
    Spline Flow & 5 & \textbf{0.65}$\pm$\small{0.01} & 12.20$\pm$\small{0.03}  & \textbf{-13.15}$\pm$\small{0.19} & -11.09$\pm$\small{0.01} & \textbf{159.47}$\pm$\small{0.02} \\
    \bottomrule
  \end{tabular}
  \end{adjustbox}
  \label{tab:uci_ablations}
\end{table}

\section{Conclusions}

In this paper we have introduced T-NAFs, a new class of neural flows that exploits transformers as an autoregressive conditioner mechanism that generates the parameters of invertible transformations. We have showcased the performance of the model on a variety of real-world density estimation problems from the classic UCI benchmarks. Overall the results have demonstrated that T-NAFs offer a powerful combination of efficiency, scalability, and flexibility, outperforming existing models like B-NAFs and NAFs in various scenarios. The innovative use of transformer architectures within T-NAFs not only addresses the scalability and training stability issues inherent in traditional autoregressive models but also paves the way for more sophisticated and nuanced modeling of complex data distributions. Future work can explore further optimizations and applications of T-NAFs, potentially expanding their utility in broader machine learning and data science contexts.

\textbf{Limitations} There are two main limitations of T-NAF that are worth discussing. (i) While T-NAF is efficient in terms of parameters, it can be expensive in terms of FLOPs due to the quadratic cost of the attention mechanism. However, this problem can be tackled by using recently proposed methods like FlashAttention \citealt{dao2022flashattention} and PagedAttention \citealt{kwon2023efficient}. (ii) T-NAF has a more complex architecture than NAF and B-NAF, therefore there are more factors that may play a role in the performance of the model.

\subsubsection*{Acknowledgments}
Funding in direct support of this work: Massimiliano Patacchiola, Aliaksandra Shysheya, and Richard E. Turner are supported by an EPSRC Prosperity Partnership EP/T005386/1 between the EPSRC, Microsoft Research and the University of Cambridge.

\bibliography{iclr2024_conference}
\bibliographystyle{iclr2024_conference}

%\appendix
%\section{Appendix}

\end{document}